# Ensemble Transfer Learning of Elastography and B-mode Breast Ultrasound Images

Sampa Misra[1], Seungwan Jeon[2], Ravi Managuli[3], Seiyon Lee[1], Gyuwon Kim[2], Seungchul Lee[2], Richard G Barr[4*], and Chulhong Kim[1,2*], Senior Member, IEEE

*Abstract*— Computer-aided detection (CAD) of benign and malignant breast lesions becomes increasingly essential in breast ultrasound (US) imaging. The CAD systems rely on imaging features identified by the medical experts for their performance, whereas deep learning (DL) methods automatically extract features from the data. The challenge of the DL is the insufficiency of breast US images available to train the DL models.  Here, we present an ensemble transfer learning model to classify benign and malignant breast tumors using B-mode breast US (B-US) and strain elastography breast US (SE-US) images. This model combines semantic features from AlexNet & ResNet models to classify benign from malignant tumors. We use both B-US and SE-US images to train the model and classify the tumors. We retrospectively gathered 85 patients' data, with 42 benign and 43 malignant cases confirmed with the biopsy. Each patient had multiple B-US and their corresponding SE-US images, and the total dataset contained 261 B-US images and 261 SE-US images. Experimental results show that our ensemble model achieves a sensitivity of 88.89% and specificity of 91.10%. These diagnostic performances of the proposed method are equivalent to or better than manual identification. Thus, our proposed ensemble learning method would facilitate detecting early breast cancer, reliably improving patient care.

## I. Introduction

Ultrasound imaging is a potential diagnostic tool to detect breast lesions. It has a high sensitivity of detecting breast cancer in women with high-risk dense breast tissue [1]. B-mode breast ultrasound (B-US) and breast US elastography are widely used for breast cancer detection [2, 3]. The breast US elastography is a novel imaging technique developed in the last decade, which shows soft lesions as red and hard lesions as blue (or vice versa), and that soft lesions tend to be benign and hard lesions malignant.  Two types of elastography employed are strain and shear-wave imaging. The strain elastography ultrasound (SE-US) utilizes tissue displacement for determining the soft from hard tissue, while shear-wave elastography ultrasound (SWE-US) determines the speed of the shear-wave inside the tumor. Both techniques are being used clinically with diagnostic accuracy better than standalone B-US imaging.

Misdiagnosis of breast cancer using US imaging is still problematic owing to the shortage of radiologists, human mistakes, and imaging quality. Thus, computer-aided detection (CAD) systems have significantly been investigated to assist diagnostic decisions [4]. Based on the traditional machine learning methods, many researchers have developed CAD systems to detect breast cancer [5, 6]. Even though these conventional machine learning methods showed promising results, they are time-consuming and challenging to design.

Recently, deep learning (DL) techniques are increasingly being applied to breast US imaging to help clinicians differentiate benign from malignant tumors. For B-US images, DL algorithms recognize patterns of lesions, e.g., orientation, boundary, echogenicity, to classify them using BI-RADS scoring. Among various DL neural networks, the convolutional neural network (CNN) [7] is most frequently used in the image classification task. It may not be simple to train CNN from scratch as it needs a large amount of labeled image data set. Collecting a large amount of breast US data is time-consuming and is often unavailable. Thus, transfer learning (TL) is suggested as one solution to address the insufficiency of breast US images for training the CNN model [8]. Byra et al. [9] developed a TL-based approach using a pre-trained VGG-19 model for breast mass classification after converting grayscale B-US images to red, green, and blue (RGB) images. Tanaka et al. [10] developed an ensemble TL model by combining two CNN architectures (VGG-19 and ResNet-152) to classify breast US images as benign or malignant using the average probability value. A DL method based on the point-wise gated Boltzmann and the restricted Boltzmann machine has been used in SWE-US to detect breast lesions and differentiate breast masses [11].  The CNN models are used to discriminate benign and malignant breast masses using SWE-US images [12]. Zhang et al. [13] reported that DL-based radiomics signatures derived from B-US and SWE-US have better diagnostic performance in

This work was supported by the National Research Foundation (NRF) grant (NRF-2019R1A2C2006269 and 2020M3H2A1078045) funded by Ministry of Science and ICT (MSIT), Institute of Information & communications Technology Planning & Evaluation (IITP) grant (No. 2019-0-01906, Artificial Intelligence Graduate School Program) funded by MSIT, Tech Incubator Program for Startup (TIPS) program (S2640139) funded by Ministry of Small and Medium-sized Enterprises and Startups (SMEs), and Basic Science Research Program through the NRF grant (2020R1A6A1A03047902) funded by the Ministry of Education, Republic of Korea.

 S. Misra, S. Y. Lee, and *C. Kim are with Opticho, Pohang 37673, Republic of Korea (e-mail: sampamisra.opticho@gmail.com; ro4797.opticho@gmail.com; chulhong@postech.edu).

 S. Jeon, G. Kim, S. C. Lee, and *C. Kim are with the Department of Electrical Engineering, Creative IT Engineering, Mechanical Engineering, and Medical Device Innovation Center, the Graduate School of Artificial Intelligence, Pohang University of Science and Technology, Pohang 37673, Republic of Korea (e-mail: jsw777@postech.ac.kr; gyuwonkim96@postech.ac.kr; seunglee@postech.ac.kr; chulhong@postech.edu).

 R. Managuli is with the Department of Bioengineering, University of Washington, Seattle 98195, USA (e-mail: ravim@uw.edu).

 R.G. Barr* is with Southwoods Imaging, 7623 Market Street, Youngstown, OH 44512, USA (e-mail: rgbarr@zoominternet.net).

classifying breast masses than those of quantitative SWE parameters and radiologist assessment.

To the best of our knowledge, it is yet to be reported a TL-based method using SE-US images alone or by combining with B-US images to identify malignant masses in the breast. Here, we present a new approach to classify breast US images (B-US and SE-US) using ensemble transfer learning. We combine B-US and SE-US images stack wise (B-SE-US) to extract features for better diagnosis. Here, various classification models are combined into one superior classifier for high predictive performance. In this approach, the CNN model is first trained using an ImageNet (large annotated natural images) dataset [14] and then the model is fine-tuned (optimization) using a small annotated breast US dataset. Among different CNN models, we used AlexNet by Krizhevsky et al. [14] and deep residual network (ResNet) by He et al. [15] because these models are well recognized with low false-positive with good accuracy in medical image classification [16, 17]. We have evaluated the model performance using both image-wise and patient-wise classification. The patient-wise classification accuracy was ~3% more than the image-wise classification. We also trained the model using entire breast US images as well as using only cropped images (i.e., the majority of the image occupied by lesion). The performance of the network trained using the cropped data set shows slightly better performance (e.g., the sensitivity of 88.89% and specificity of 91.10% for patient-wise) compared to the network trained using the full image (e.g., sensitivity of 82.22 % and specificity of 88.76% for patient-wise). The details of our implementation and test results are presented in the next few sections.

## II. MATERIALS AND METHODS

### A. Dataset and Data Augmentation

Eighty-five patients were retrospectively included in this IRB approved and HIIPAA complaint study (approved by Western reserve health education, Youngstown, Ohio) conducted from August of 2016 through February 2017. The patient details can be found in Supplementary Table 1 and Table 2. There were no inclusion/exclusion criteria for selecting the patient and the final diagnosis was confirmed by biopsy. Each patient had multiple B-US and their corresponding SE-US images. The study contained 261 B-US images and 261 SE-US images.

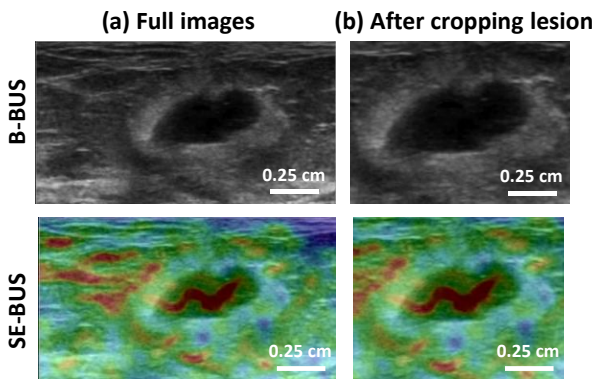

**Fig. 1** Dataset preparation for B-BUS and SE-BUS images. (a) Full images (b) Cropped images of the lesion parts.

More details on acquiring clinical strain images can be found in [18]. The datasets were acquired using Hitachi's HI VISION Ascendus system with EUP-L75 probe (linear-type probe 18–5 MHz). The dataset consists of 130 benign (*B*) and 131 malignant (*M*) JPEG images. An example of the full image and the corresponding cropped image containing only the lesion part are shown in Fig. 1a and Fig. 1b, respectively. The specific lesion location is cropped manually based on the radiologist's suggestion. After cropping, the images are known as 'after cropping lesion'.

We also applied a technique called augmentation, commonly used in DL, to expand the training dataset to reflect the lesion's different presentation, e.g., rotation, size, texture. Here, we applied random cropping and horizontal flip as an augmentation strategy. We have used online augmentation where transformation is done in mini-batches, so the model is trained with different images at each epoch.

### B. Deep Learning (DL)

The CNN-based DL model is widely used in medical image classification because of its excellent feature extractor capability [14, 19]. Generally, the architecture of CNN consists of 4 layers: convolutional layers (employed on the input images to extract features), ReLU (rectifier linear unit) layers (to use an elementwise activation function to increase non-linearity in the network), pooling layers (to perform downsampling operation by keeping only the high-level features), and FC (fully connected) layers (to connect neurons to produce the result). The accuracy of the CNN model is influenced by the size of the training data and layer design. As previously indicated, since the data size is limited in medical imaging, TL is used, where the models are pre-trained using a sizeable natural image dataset. Such TL approaches have already been used in various medical imaging applications [16, 20]. Here, we applied the TL approach to elastography imaging as well. In the TL, the network is first trained on a different broad field data set (e.g., ImageNet) before applying to a specific domain.

### C. Transfer Learning (TL)

The neural network, for differentiating benign from malignant lesions, is trained in two stages: 1) pre-training: the network is trained on a large-scale ImageNet dataset with 1000s of labels/categories and 2) fine-tuning: the pre-trained network is further trained on the breast US dataset with 2 labels (benign and malignant). The fine-tuning helps to adapt a pre-trained CNN to a different dataset by updating the pre-trained weights using backpropagation [21]. In general, low-level features (edges, corners, and curves) are learned in the initial layers of the model, and the high-level features (specific to breast images) are learned in the final layers of the model. Thus, during fine-tuning, the first few layers of the pre-trained model are kept frozen, while the rest of the layers are updated.

**AlexNet:** The structure of the AlexNet model used in this study is shown in Fig. 2. This architecture was developed by Alex Kriszhevsky [14] and achieved the highest classification accuracy at the ImageNet Large Scale Visual Recognition Challenge (ILSVRC) in 2012 [14]. The deep structure contains eight main layers, out of which the first five are convolutional layers, and the rest three are FC layers. There are overlapping max-pooling layers after the first, second, and fifth convolutional layers to reduce overfitting problems. A ReLu activation is applied to each convolutional layer, which accelerates the training with high accuracy. The fifth convolutional layer (after max and avg pooling) is linked to the FC layers. We have used the trial-and-error method to determine frozen and re-trained layers. Based on the result, the last three layers of the AlexNet model are fine-tuned using our breast dataset, where the number of classes is two (benign and malignant) instead of 1000s of categories of ImageNet data.

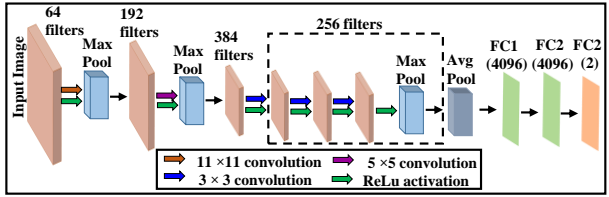

**Fig. 2** AlexNet architecture used in the proposed method.

**ResNet:** Deep residual network (ResNet) is used for image classification to take advantage of low training error [15]. The classification performance and feature extraction using ImageNet dataset of different ResNet models like ResNet-152, ResNet-101, ResNet-50, ResNet-34, and ResNet-18 (the value indicates the number of network layers) are better than other CNN models [15]. We choose ResNet-18 because of its shallow architecture and faster training without compromising performance. The ResNet-18's network structure is presented in Fig. 3. It comprises one 7×7 convolutional layer (followed by batch normalization layer and ReLu layer), 5 residual blocks, 2 pooling layer, and one FC layer. Among the 5 residual blocks, 2 are regular residual blocks (Res block1), and 3 are residual blocks with 1×1 convolution (Res block2), which are used for changing the number of channels. The block consists of two 3×3 convolutional layers, 2 batch normalization layer, and one ReLU layer. In the case of ResNet also, the trial-and-error method is used to determine frozen and re-trained layers. Based on the result, the first four layers were frozen. Classification accuracy of this model is enhanced with faster training because the network is protected from vanishing gradient problem due to the availability of shortcut connection, and it avoids negative outcomes while increasing network depth [15]. The vanishing gradient problem is discussed in the supplementary file.

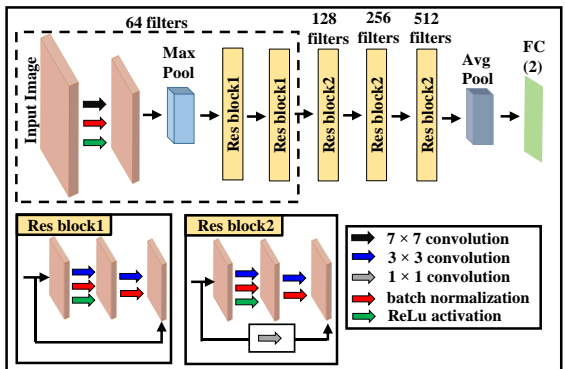

**Fig. 3** The network structure of ResNet-18. The regular ResNet block is denoted as Res block1 and ResNet block with 1x1 convolution is denoted as Res block2. ReLu stands for rectified linear unit activation. The fully connected (FC) layer is with 2 outputs (benign and malignant).

### D. Ensemble Learning

The overview of our proposed **ensemble** model is shown in Fig. 4. The p-trained AlexNet and ResNet models are used in this investigation for feature extraction and then these features are used to classify the images using a softmax classifier.

**Ensemble Design:** The step-by-step process to combine B-US and SE-US images stack wise to make B-SE-US images and the proposed **ensemble** model from the AlexNet & ResNet models trained using B-SE-US images are described here.

i. The steps to combine B-US and SE-US images stack wise are shown in Fig. 4 and specified as follows:
   1. Resize B-US breast cancer images (*H, W,* 1) (*H* and *W* are the height and width of the images, respectively, and 1 is the number of channels (gray) for grayscale images) as different images had different sizes. The images were resized to 224 × 224 pixels.
   2. Resize SE-US breast cancer images (*H, W,* 3) (3 is the number of channels (red, green, and blue) for color images). The images were resized to 224 × 224 pixels.
   3. Combine B-US and SE-US breast cancer images (B-SE-US) stack wise (H, W, 4) (4 is the number of channels (gray, red, green, and blue) for the combined grayscale and color images). The B-US and SE-US images are not co-aligned to the same spatial coordinates. The pooling layers help to align the images, eventually.

ii. The steps to ensemble AlexNet and ResNet models are illustrated in Fig. 4 and enumerated as follows:
   1. **Load AlexNet and ResNet models** – These are neural networks already trained using the ImageNet dataset. Initialize both neural networks with pre-trained weights instead of random weights to reduce training time. The pre-trained weights using the ImageNet dataset for both of these models are already available in PyTorch [22].
   2. **Fine-tune AlexNet and ResNet models** – To train models to **breast** US image dataset (smaller than ImageNet dataset), few initial layers of the two models are frozen (fixed at initial weights), and only the higher layers (from last frozen layer's output to respective classification layers) are re-trained using B-SE-US images. In Figure 4, Step 2, gray layers imply fixed weights, and colored layers are trained.
   3. **Remove classification (softmax) layer** – As the fine-tuned AlexNet and ResNet models are to be combined to create the ensemble model, the intermediate output of their penultimate layer is exposed.
   4. **Concatenation and add classification layer** – After Step 3, we have feature maps for each model optimized using B-SE-US images. They are concatenated to obtain a super feature set. A common classification (softmax) layer with randomly initialized connections is used to obtain the final classification (benign/malignant) of the images.
   5. **Fine-tune new ensemble model** – Similar to step (ii), the weights of the constituent models are frozen, and the classification layer of the model is re-trained to fit the B-SE-US images.

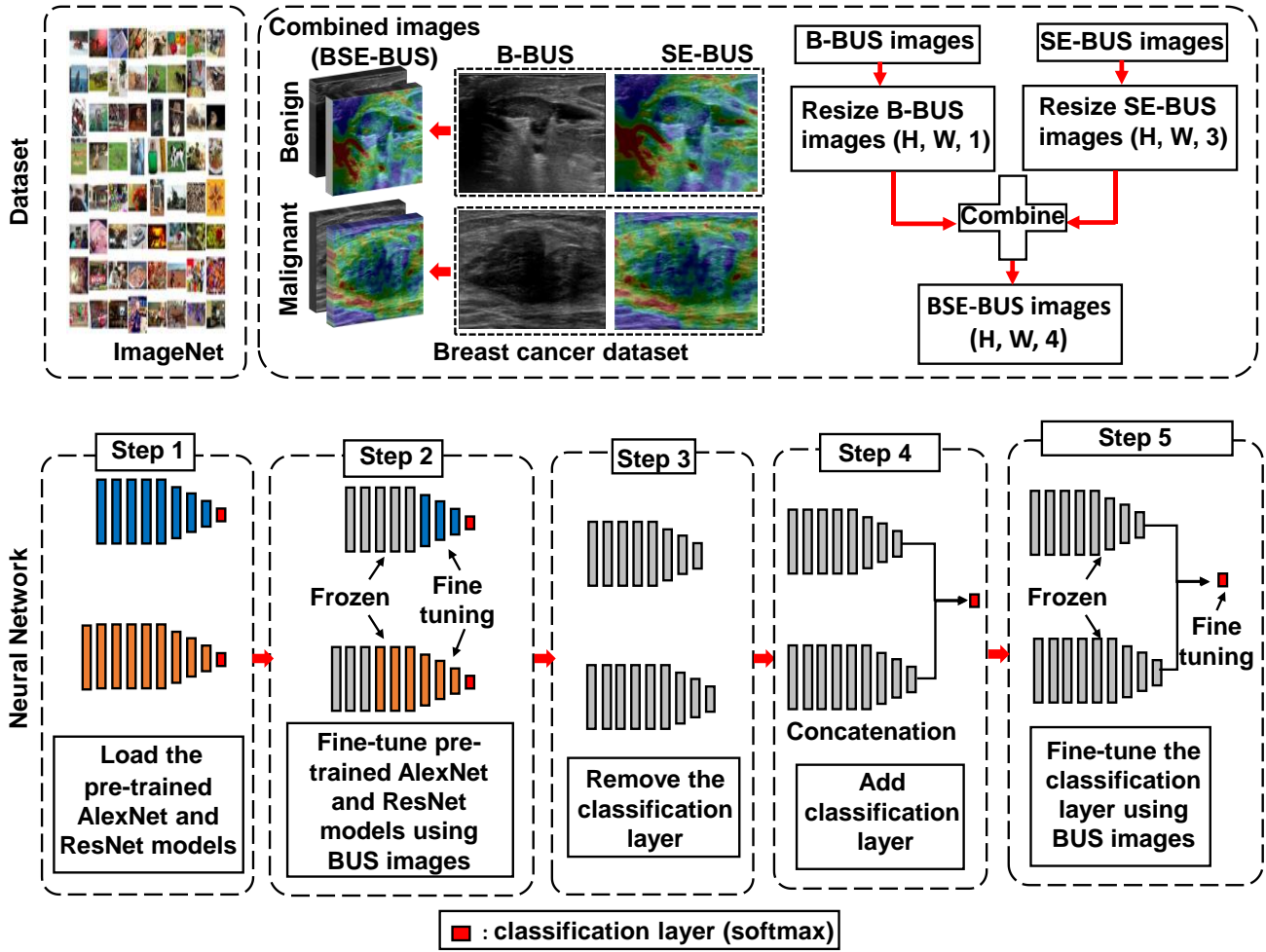

**Fig. 4** Schematic illustration of the methods employed in this proposed study. The ImageNet data set is used to train the model and stack wise combined B-BUS and SE-BUS breast cancer images are used for fine-tuning. First, ResNet (cyan color) and AlexNet (blue color) models are trained using the ImageNet dataset. Next, some high layers of ResNet and AlexNet are fine-tuned using a combined breast cancer image dataset to distinguish between the benign and the malignant while keeping the other layers frozen. The proposed classifier is obtained by combining the respective ResNet and AlexNet classifier. BUS, breast ultrasound; B-BUS, b-mode breast ultrasound; SE-BUS, strain elastography ultrasound; BSE-BUS, combined b-mode and strain elastography ultrasound; H and W are the height and width of the images. 1, 3, & 4 are the number of channels (gray) for grayscale images, color images, and combined grayscale & color images.

The AlexNet and ResNet models contain 4096 and 512 features, respectively, and the proposed model contains a total of 4608 features after concatenation.

### E. Experimental Setup

We used PyTorch as a back-end language for the DL models implementation and ran our model on a server equipped with a super GPU, GeForce RTX 2060. For training, we employed the most popular cross-entropy as loss function and Adam optimizer [23] as optimization function with a learning rate of 0.0001. The cross-entropy loss function minimizes the differences between the predicted and actual probability distributions. Adam optimizer is computationally efficient and is also easy to implement. An early stopping criterion (if the validation loss did not change in 200 consecutive epochs, then stop training) was also implemented, and the weights of the best epoch were restored from stored values as the best weights are saved in memory. These parameters were fixed for all the CNN models used in the manuscript since we wanted to compare the performance of the proposed model with existing models for the same dataset.

## III. RESULTS AND DISCUSSION

Out of 85 patients, 65 were assigned as training patients, and 20 patients were assigned as testing patients. This corresponds to the 80:20 ratio between training and testing patients. All the images belonging to the training patient were used for training, and all the images belonging to the testing patient were used for testing. They were never mixed. The data were divided patient-wise because often, multiple SE-US images acquired for the same patient are not diagnosed consistently. For example, 3 images acquired for a patient (PT# 208) are shown in Supplementary Fig. 1. The lesion in two images (Supplementary Figs. 1a and 1b) were diagnosed as benign, but the last image was diagnosed as malignant (Supplementary Fig. 1c). Such variation in outcome arises due to the variations that exist in acquiring the elastography images [24]. We have employed five-fold cross-validation to validate the performance of the proposed model. The training data in-turn was randomly divided into five parts. From these five parts, four parts were assigned as the training data and the remaining part as the validation data. Again, as stated before, all images acquired from a specific patient exclusively belong to either test, train, or validation.

**Table 1.** The image-wise and patient-wise cross-validated classification performance (mean ±SD) results (accuracy, precision, specificity, and sensitivity) when the network was trained with only B-BUS images, SE-BUS images, and combined (BSE-BUS) images. In the third case, the dataset consists of stacks of B-BUS images and SE-BUS images. The red box in the table denotes the highest performance values.

| | | Full image | | | | | after cropping lesion | | | | |
|---|---|---|---|---|---|---|---|---|---|---|---|
| | | Accuracy (%) | Precision (%) | Specificity (%) | Sensitivity (%) | F1-score (%) | Accuracy (%) | Precision (%) | Specificity (%) | Sensitivity (%) | F1-Score (%) |
| | | **Patient-Wise** | | | | | | | | | |
| B-BUS | AlexNet | 66.67±3.08 | 66.73±3.80 | 66.67±3.02 | 66.67±3.97 | 66.41±2.91 | 71.11±2.22 | 71.73±4.48 | 71.11±3.88 | 71.11±5.44 | 71.09±1.31 |
| | ResNet | 66.66±4.96 | 67.78±3.64 | 68.89±3.31 | 64.48±3.31 | 65.74±2.50 | 71.11±2.22 | 72.22±3.43 | 73.33±3.44 | 68.89±4.44 | 70.42±2.22 |
| | Ensemble | 72.22±3.51 | 72.89±3.99 | 73.34±3.44 | 71.11±3.44 | 71.86±3.70 | 77.78±3.51 | 77.16±3.97 | 75.56±4.31 | 80.00±4.44 | 78.32±2.86 |
| SE-BUS | AlexNet | 75.56±2.72 | 75.66±1.92 | 75.56±4.44 | 75.56±3.31 | 75.35±3.96 | 81.11±2.71 | 81.10±3.73 | 80.00±4.31 | 82.22±5.44 | 81.33±2.22 |
| | ResNet | 75.55±4.44 | 77.06±5.79 | 77.78±3.02 | 73.33±5.44 | 74.99±4.52 | 81.11±2.71 | 81.10±3.73 | 80.00±4.31 | 82.22±5.44 | 81.33±2.22 |
| | Ensemble | 78.89±2.22 | 78.22±0.22 | 77.78±0.00 | 80.00±4.44 | 79.07±2.57 | 85.55±2.71 | 86.55±3.33 | 86.66±4.42 | 84.44±5.44 | 85.33±2.97 |
| BSE-BUS | AlexNet | 81.11±2.71 | 82.11±4.47 | 82.22±5.44 | 80.00±4.44 | 80.89±2.63 | 84.44±4.15 | 86.10±4.20 | 86.66±4.42 | 82.22±5.53 | 84.05±4.28 |
| | ResNet | 81.94±2.40 | 83.19±4.37 | 83.33±5.55 | 80.56±4.81 | 81.67±2.37 | 84.44±2.22 | 85.97±3.93 | 84.44±5.44 | 84.44±5.44 | 84.40±2.39 |
| | Ensemble | **85.55±2.72** | **88.76±1.42** | **88.89±2.02** | **82.22±1.44** | **85.06±2.68** | **90.00±2.15** | **91.05±1.54** | **91.10±1.44** | **88.89±2.02** | **89.79±4.47** |
| | | **Image-wise** | | | | | | | | | |
| B-BUS | AlexNet | 65.7±3.97 | 64.03±3.90 | 65.51±4.87 | 65.92±3.18 | 64.82±4.88 | 68.21±4.43 | 68.29±3.79 | 71.71±3.03 | 64.44±3.02 | 66.16±4.51 |
| | ResNet | 66.78±4.73 | 66.92±5.08 | 71.03±7.10 | 62.22±3.22 | 64.19±5.53 | 67.85±3.19 | 67.52±3.78 | 71.03±4.14 | 64.44±2.96 | 65.92±3.12 |
| | Ensemble | 71.06±4.57 | 71.78±5.89 | 75.16±7.03 | 66.66±3.19 | 68.92±4.81 | 73.56±4.97 | 74.46±3.44 | 76.54±3.09 | 70.37±3.73 | 66.16±4.51 |
| SE-BUS | AlexNet | 75.71±1.82 | 75.69±1.96 | 77.92±3.51 | 73.33±3.92 | 74.32±2.80 | 81.78±1.33 | 80.11±4.58 | 79.99±3.68 | 83.68±3.54 | 81.56±1.16 |
| | ResNet | 74.99±1.12 | 74.88±1.96 | 77.25±2.77 | 72.59±1.81 | 73.68±1.00 | 80.71±3.81 | 80.68±8.02 | 80.68±3.36 | 80.74±3.62 | 80.31±2.57 |
| | Ensemble | 77.49±1.42 | 76.26±2.67 | 77.24±4.14 | 77.77±4.05 | 76.89±1.54 | 83.92±1.12 | 84.52±4.12 | 85.51±3.51 | 82.22±4.31 | 83.13±0.87 |
| BSE-BUS | AlexNet | 81.42±3.11 | 82.92±6.93 | 84.13±7.74 | 78.51±3.62 | 80.38±2.53 | 84.63±1.82 | 87.26±3.12 | 88.96±3.38 | 79.99±3.77 | 83.37±2.04 |
| | ResNet | 79.90±2.92 | 80.27±4.58 | 81.89±5.09 | 77.77±4.53 | 78.86±2.96 | 81.42±2.41 | 83.71±5.28 | 83.44±4.31 | 79.26±1.81 | 80.52±1.69 |
| | Ensemble (voting) | 82.12±2.41 | 83.92±5.13 | 85.13±6.31 | 78.71±3.12 | 82.38±2.54 | 86.63±2.10 | 86.26±3.13 | 86.96±2.56 | 82.99±2.76 | 84.37±1.98 |
| | Ensemble | **84.28±2.37** | **86.24±5.20** | **87.58±5.16** | **80.73±4.31** | **83.20±2.40** | **87.48±2.26** | **88.49±2.15** | **89.65±2.18** | **85.18±4.05** | **86.75±2.63** |

For testing, the dataset is divided into image-wise and patient-wise (See Supplementary Fig. 2). Although one patient is classified in a particular class based on majority voting [25], in case of a tie, importance is given to the malignant case. Therefore, the benign (*B*) or malignancy (*M*) of a tumor in a patient is determined by:

$$\text{Patient} = \begin{cases} B, \text{ if } B_P > N_P - B_P \\ M, \text{ Otherwise} \end{cases} \quad (1)$$

where $N_P$ is the total number of images of the patient $P$ (since multiple images are typically acquired for each patient) and $B_P$ is the number of images predicted as benign for the patient.

The image-wise and patient-wise cross-validated performance for results (accuracy, precision, specificity, sensitivity, and F1-score) for AlexNet, ResNet, and ensemble models are shown in Table 1 for B-US, SE-US, and B-SE-US images. In B-SE-US, networks were trained using stacked B-US and SE-US images. The network performance is provided with two sets of data: (1) networks trained with the full images (e.g., Fig. 1a), (2) networks trained with cropped images containing majority lesion (Fig. 1b). In the next few subsections, we present our results and also provide appropriate discussion.

*Patient-wise vs. image-vise comparison:*

The classification accuracy is ~90% using the patient-wise compared to ~87% using the image-wise. For the image-wise case, the classification was performed on a pool of 56 test images, where the characteristics of every individual image were taken into consideration. For patient-wise, we consider the characteristics of the majority of the images pertaining to the patient to assign the diagnosis (benign or malignant). Hence in the latter case, we eliminate the probability of misclassification through the majority wins rule. For example, if a patient has 4 images and the model misclassifies 1 image among them, the classification accuracy will be 75% in the image-wise classification, but this accuracy value will be 100% in the patient-wise classification, because of the majority-win rule.

**Table 2.** The patient-wise cross-validated classification performance (mean ± SD) of existing methods (i.e., MSR, ASR, DenseNet, VGG, SqueezeNet, AlexNet, and ResNet) and our proposed method (ensemble of AlexNet and ResNet) using the BSE-BUS images. The red box denotes the proposed method and its performance values. MSR, manual strain ratio; ASR, assist strain ratio.

|  | Full image | | | | | after cropping lesion | | | | |
|---|---|---|---|---|---|---|---|---|---|---|
|  | Accuracy (%) | Precision (%) | Specificity (%) | Sensitivity (%) | F1-score (%) | Accuracy (%) | Precision (%) | Specificity (%) | Sensitivity (%) | F1-Score (%) |
| **MSR** | 76.78 | 70.58 | 88.88 | 65.52 | 78.67 |  |  |  |  |  |
| **ASR** | 80.35 | 76.66 | 85.18 | 75.86 | 81.09 |  |  |  |  |  |
| **VGG** | 72.22±7.02 | 73.14±8.13 | 73.33±8.88 | 71.11±8.88 | 71.84±6.97 | 75.55±4.44 | 77.64±7.56 | 77.78±9.93 | 73.33±5.44 | 75.06±3.92 |
| **SqueezeNet** | 75.56±4.44 | 78.73±6.99 | 80.00±8.31 | 71.11±5.44 | 74.44±4.08 | 76.67±2.22 | 79.39±5.86 | 80.00±8.31 | 73.34±5.44 | 75.85±1.69 |
| **DenseNet** | 77.78±3.51 | 80.75±4.91 | 82.22±5.44 | 73.33±5.44 | 76.70±3.85 | 81.11±2.71 | 83.69±4.04 | 84.45±5.44 | 77.78±7.02 | 80.33±3.40 |
| **AlexNet** | 81.11±2.71 | 82.11±4.47 | 82.22±5.44 | 80.00±4.44 | 80.89±2.63 | 84.44±4.15 | 86.10±4.20 | 86.66±4.42 | 82.22±5.53 | 84.05±4.28 |
| **ResNet** | 81.94±2.40 | 83.19±4.37 | 83.33±5.55 | 80.56±4.81 | 81.67±2.37 | 84.44±2.22 | 85.97±3.93 | 84.44±5.44 | 84.44±5.44 | 84.40±2.39 |
| **Ensemble** | 85.55±2.72 | 88.76±1.42 | 88.89±2.02 | 82.22±1.44 | 85.06±2.68 | 90.00±2.15 | 91.05±1.54 | 91.10±1.44 | 88.89±2.02 | 89.79±4.47 |

Banerjee et al. [25] reported that the patient-wise classification improved the accuracy by ~10% compared to the image-wise classification. They considered random image separation from the pool of total images for the training and testing purposes without considering the impact of the same patient images in both pools. Hence, there always remains a possibility of data leakage (i.e., the same patients' images may consider for training and testing). However, during the patient-wise classification, they completely isolated the test set from the training one. Thus, large improvements in accuracy (e.g., ~10%) were justified. However, in our case, the accuracy enhancement is ~3% with the patient-wise classification. We completely isolated the training and testing data for both image-wise and patient-wise classifications. We divided our dataset into training, validation, and test sets in such a way that the patient images used to build the test set was not used for the training and validation sets. Therefore, the accuracy enhancement after the patient-wise classification is only attributable to the reduction of misclassification through majority voting.

We have compared our proposed method to the majority voting ensemble method where ResNet and AlexNet models are trained and then outputs from these models are utilized by making a vote. It is worth noting that the performance of our proposed ensemble model is better than ensemble model with majority voting.

*Cropped vs. full image comparison:*

From Table 1, it can be seen that the performance of the network trained using the cropped data set shows slightly better performance compared to the network trained using the full image. The patient-wise accuracies improved from 66.67% to 71.11% when the AlexNet model was trained using cropped B-US images instead of full B-US images. Further, these accuracies improved from 71.11% to 84.44% when the AlexNet model was trained using cropped B-SE-US images instead of cropped B-US images. Sensitivity and specificity increased from 71.11% to 82.22% (sensitivity) and 71.11% to 86.66% (specificity) when the AlexNet model was trained using B-SE-US images compared to B-US images. These values are 84.44% (sensitivity) and 84.44% (specificity) when the ResNet model was trained using B-SE-US images. The performance of the proposed ensemble model improves to a great extent in terms of sensitivity (88.89%) and specificity (91.10%) using the same B-SE-US images. It is worth mentioning that the classification performance of the model trained using B-SE-US images (i.e., B-US and SE-US) is better than that of the model trained using B-US or SE-US images alone. The classification performance of the proposed ensemble model is superior to the individual AlexNet, or ResNet model trained using all B-US, SE-US, and B-SE-US images. Finally, these values are further improved by training the model using the cropped images of the lesion parts instead of full images.

*Comparing against existing models:*

Since strain imaging is qualitative, several techniques have been developed to quantify these images, e.g., Tsukuba score [26], Fat-to-Lesion ratio (FLR) [27], Elastography over BW size ratio (E/B ratio) [28]. Barr & Managuli [18] compared the diagnostic performance for assist strain ratio (ASR) and manual strain ratio (MSR) using the same full breast US images, which are used in this study. We also compared our proposed ensemble model with other well-known CNN models like DenseNet, VGG, and SqueezeNet [29] models using the full images as well as cropped images. The cross-validated performance metrics (i.e., accuracy, precision, specificity, sensitivity, and F1-score) comparing the proposed model with other existing methods for both full and cropped images are shown in Table 2. The performance metrics are calculated using the B-SE-US images for all the models since the performance is better using B-SE-US images as compared to B-US images and SE-US images. Among the existing CNN models, the performance of the VGG model is most inferior and the performance of the AlexNet and the ResNet models is superior. It is worth noting that the performance of the proposed model is superior compared to the other existing methods for the same dataset.

*Global patient recognition:*

**Table 3.** Patient recognition rate for the network trained with only B-BUS, only SE-BUS, and BSE-BUS full and cropped images. The red boxes indicate the performance of the proposed method.

|  |  | Patient recognition rate (%) | | |
|---|---|---|---|---|
|  |  | AlexNet | ResNet | Ensemble |
| Full images | B-BUS | 67.59 | 66.67 | 72.78 |
|  | SE-BUS | 74.63 | 73.14 | 76.76 |
|  | BSE-BUS | 80.92 | 78.61 | 81.67 |
| after cropping lesion | B-BUS | 68.98 | 68.42 | 73.24 |
|  | SE-BUS | 81.67 | 80.09 | 81.94 |
|  | BSE-BUS | 85.00 | 82.74 | 86.39 |

The global patient recognition rate when multiple images are present for the same patient is determined as [30]:

$$\frac{1}{N} \sum \frac{N_{cor}}{N_P} \qquad (2)$$

where $N$ is the total number of patients, $N_P$ is the number of images of patient $P$ and $N_{cor}$ is the number of images correctly classified as benign or malignant for that specific patient. The global patient recognition rate for only B-US, only SE-US, and both B-SE-US considering both full and cropped images are shown in Table 3. The global patient recognition rate for the AlexNet and ResNet models is improved from 68.98% to 85.00% and 68.42% to 82.74%, respectively when the models were trained using cropped B-SE-US images instead of cropped B-US images. The ensemble model further improves this value by up to 86.39%.

*Statistical analysis:*

The images are classified based on a higher predictive probability value (PPV) calculated from the softmax classifier. The proposed ensemble model is statistically (*t*-statistics [31]) validated with the ResNet or AlexNet models independently using the distributions of PPV values that related to the actual class labels. Fig. 5(a) shows the statistical analysis of the PPV values for ResNet, AlexNet, and the ensemble models using the cropped B-US, SE-US, and B-SE-US images. The related *p*-values for the different models (ensemble & ResNet and ensemble & AlexNet) and the different image types (B-US & SE-US, B-US & B-SE-US, and SE-US & B-SE-US) are shown in Fig 5(b) and 5(c), respectively. The mean PPV values for the proposed ensemble model using B-SE-US images are superior compared to all methods, and it is statistically validated with a 95% confidence level for B-US, SE-US, and B-SE-US images.

*Interpretability of the model:*

Recently, the use of DL in medical imaging emphasizes the interpretability of the model to determine the reliability of the proposed system. Specifically, for CNN, numerous methods have been proposed to understand the black-box working principle of the model [32, 33]. In this study, Grad-CAM (Gradient-weighted Class Activation Mapping) [34], a generalized algorithm of CAM (Class Activation Mapping) [35], has been utilized to understand and interpret the working principles of our proposed ensemble network. The main concept of the Grad-CAM is to linearly combine the extracted features from the final convolutional layer and up to the sample via interpolation to the original image size to obtain a heatmap. The heatmap would indicate which parts of the image were referenced by CNN for classification. In our ensemble model, two different channels exist for feature extraction, as shown in Fig 4: one from the AlexNet and the other from the ResNet. For each extractor channel, weights for individual feature maps are computed using gradient information backflowing into the last convolutional layer of the feature extractor channels. Before linearly combining all feature maps, the feature maps from the AlexNet extractor are resized to match the feature maps from the ResNet extractor. After the linear combination, the nonlinear activation function of ReLU is applied to only emphasize the features that have a positive influence on the outcome. An example of benign and malignant images and their corresponding Grad-CAM analyses are shown for both full

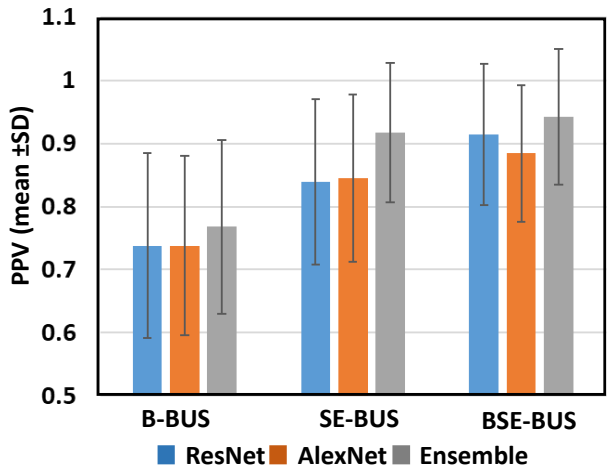

| (b) | Ensemble | | |
|---|---|---|---|
|  | B-BUS | SE-BUS | BSE-BUS |
| ResNet | .02098 | < .00001 | .00229 |
| AlexNet | .00563 | < .00001 | < .00001 |

| (c) | B-BUS & SE-BUS | B-BUS & BSE-BUS | SE-BUS & BSE-BUS |
|---|---|---|---|
| ResNet | < .00001 | < .00001 | < .00001 |
| AlexNet | < .00001 | < .00001 | .00025 |
| Ensemble | < .00001 | < .00001 | .00206 |

**Fig. 5** (a) Statistical analyses of the predictive probability value (PPV). (b) *p*-values for the different models (ensemble & ResNet and ensemble & AlexNet) for B-BUS, SE-BUS, and BSE-BUS images, and (c) *p*-values for the different sets of cropped images (B-BUS & SE-BUS, B-BUS & BSE-BUS, and SE-BUS & BSE-BUS) for ResNet, AlexNet, and ensemble models.

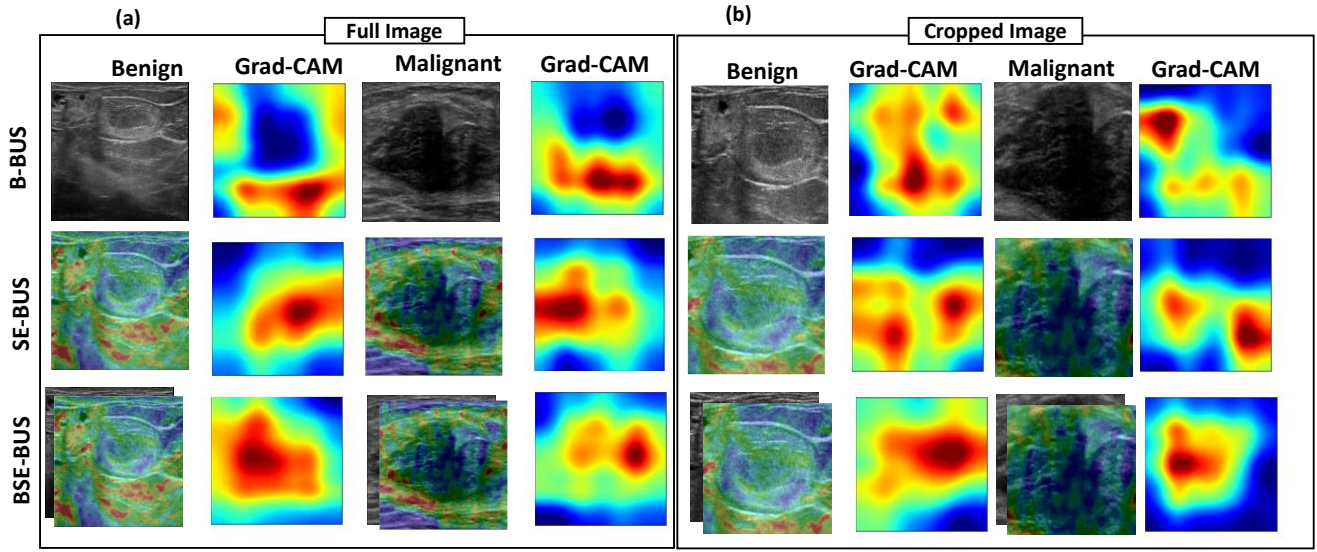

**Fig. 6** Examples of benign and malignant B-BUS, SE-BUS, and BSE-BUS images and their corresponding Grad-CAM analyses for (a) the full images and (b) cropped images.

images and corresponding cropped images in Fig. 6(a) and Fig. 6(b), respectively. The images were misclassified by the models (ensemble of AlexNet & ResNet models) when trained using only the B-US or SE-US images. However, they were correctly classified by the ensemble model, which was trained using B-SE-US images. The Grad-CAM result also showed that the ensemble model trained using the combined images (B-SE-US) identify the lesion position more accurately when compared to other ensemble models, which were trained using only the B-US or SE-US images.

The reasons for high performance are two-fold: 1) a combination of the B-US and SE-US images encoding both structural and functional information and 2) ensemble of the different models offset the weakness of the individual models. The ensemble of the architectures can be exceedingly helpful in obtaining various features. Herein, we observe that the ensemble model appears to be far prudent in accurately classifying images, which are often misclassified by individual models. The result indicates that the ensemble model influenced the strengths of the individual models and was thus capable of classifying images that would have been misclassified by individual models. Furthermore, the fine-tuning of the CNNs in our ensemble enables us to extract image features that are more relevant to the breast images being classified.

*Misclassification:*

Although the proposed method is highly efficient, it misclassifies some of the patients. Table 4 lists the pathological features of the misclassified patients (benign patients judged as malignant patients and vice versa) by the proposed ensemble model for 3-fold or more in 5-fold cross-validation. The pathological diagnosis of the misclassified cases was fibrocystic and IDC. We analyze the following aspects from misclassified patients:

1) The strain ratio of the patient (PT# 284, true class benign) is 3.3. The patient is also predicted as malignant using the strain value [18].

2) The mean PPV values are ~0.91 for the benign class and ~0.89 for the malignant class using the correctly classified patients. However, the PPV values for the misclassified benign patient (PT# 284) and malignant patient (PT# 229) are ~0.46

**Table 4.** Pathological features of misclassified patients.

| Patient No. | Histological Type | Predicted Class | No. of times Misdiagnosis in 5-fold |
|---|---|---|---|
| 284 | fibrocystic | Malignant | 3 |
| 229 | IDC | Benign | 4 |

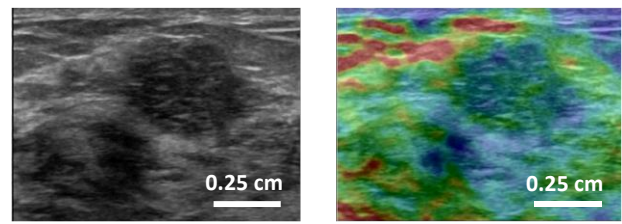

**Fig. 7** Example of misclassified malignant image. The proposed method predicts as benign.

and ~0.23, respectively.

3) The misclassified patient labeled as malignant (PT# 229) is shown in Fig. 7. The center of the lesion is shown in blue, while the peripheral areas in green. So, the score of this image is 3, according to Tsukuba Elasticity Score (TES) [36], which is particularly unclear but more likely benign.

If the number of mages in the data set is increased to train or more such images were used for training the model, the performance of our algorithm could also improve further. The proposed method correctly classifies images that are even misclassified by ASR or MSR method. For example, a patient (PT# 207) labeled as benign is shown in Fig. 8. The patient is predicted as malignant using the strain value as the strain ratio of the patient is 3.8 [18]. However, our model correctly classifies the patient as malignant.

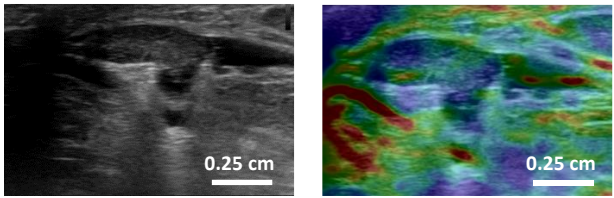

**Fig. 8** Example of correctly classified benign image.

*Limitations:*

We used cross-domain data set to TL, which implies, the data set we used for training is from a different domain (i.e., ImageNet) compared to the eventual application (medical). Although using cross-domain TL has gives an excellent performance, the models pre-trained on natural images may not be the optimal choice in the medical imaging domain. Many studies have discussed using the same domain data for TL to improve the performance. For example, ultrasound images of other body parts can be used to train the neural network. Such a pre-trained network can then be used as the initial network for breast classification.

Also, we used the same fine-tuning parameters for both ResNet and AlexNet models to demonstrate the impact of the ensemble and did not implement any parameter optimizations method. Parameter optimization for each model using cross-validation may give further improved results.

*Future work*

The performance results suggest that the proposed ensemble CNN model has advantages as a second reviewer to support the diagnostic decisions of human reviewers and provide opportunities to compare decisions. The proposed model can be integrated with existing CADx systems to provide robust breast cancer classification. In future work, it would be worthwhile to evaluate our work with dedicated detection algorithms for breast lesion detection and classification as an end-to-end CAD solution. In the future, we will check whether the performance of the proposed method can be further improved by adapting the fine-tuning system (e.g., changes to the loss function).

Our method can also be extended by integrating other CNN architectures, such as GoogLeNet. In the future, we will combine different CNN models. We also intend to provide our model available publicly for other researchers to use or to improve the model.

## IV. CONCLUSION

In this manuscript, we develop a TL-based CAD system by combining the AlexNet and ResNet models to classify the breast masses as benign or malignant using clinical B-US and SE-US images. The proposed method used two fine-tuned CNN models (AlexNet & ResNet) to extract features that were able to capture the diverse features present in the ultrasound image data. The experimental results conducted here showed that the proposed ensemble model using both B-mode and strain images could classify the majority of the images correctly and hence achieve the best performance, with high sensitivity and specificity. In the near future, we plan to use the model for a prospective study to determine the sensitivity and specificity in the real-world scenario.

ACKNOWLEDGMENT

S. Misra, S. Y. Lee, and C. Kim have financial interests in OPTICHO, which supported this work.

**Supplementary Information for:**

# Ensemble Transfer Learning of Elastography and B-mode Breast Ultrasound Images

Sampa Misra, Seungwan Jeon, Ravi Managuli, Seiyon Lee, Gyuwon Kim, Seungchul Lee, Richard G Barr*, and Chulhong Kim*

**\*Corresponding author:** Chulhong Kim (chulhong@postech.edu) and Richard G. Barr (rgbarr@zoominternet.net)

S. Misra, S. Y. Lee, and C. Kim* are with Opticho, Pohang 37673, Republic of Korea;
S. Jeon, G. Kim, S. C. Lee, and C. Kim* are with the Department of Electrical Engineering, Creative IT Engineering, Mechanical Engineering, and Medical Device Innovation Center, the Graduate School of Artificial Intelligence, Pohang University of Science and Technology, Pohang 37673, Republic of Korea;
R. Managuli is with the Department of Bioengineering, University of Washington, Seattle 98195, USA;
R.G. Barr* is with Southwoods Imaging, 7623 Market Street, Youngstown, OH 44512, USA.

**Table of contents**

**Supplementary Methods**

**Supplementary Figures**

**Supplementary Tables**

**Supplementary Method 1. The Vanishing Gradient problem and ResNet**

Theoretically, if the number of layers of the neural network (NN) is increased, performance will be improved. However, in practice, if more layers using certain activation are added to the NN, it will be difficult o train, as the gradients of the loss function approach zero. This issue is known as the vanishing gradient problem. The activation function like sigmoid squishes a large input space into a small space. Therefore, a large change in the input of the activation function will cause a small change in the output.

ResNet is one of the solutions to this problem as it provides residual or skips connection that connects layers further behind the preceding layer to a given layer. The Resnet model avoids this problem by constructing ensembles of many short networks together.

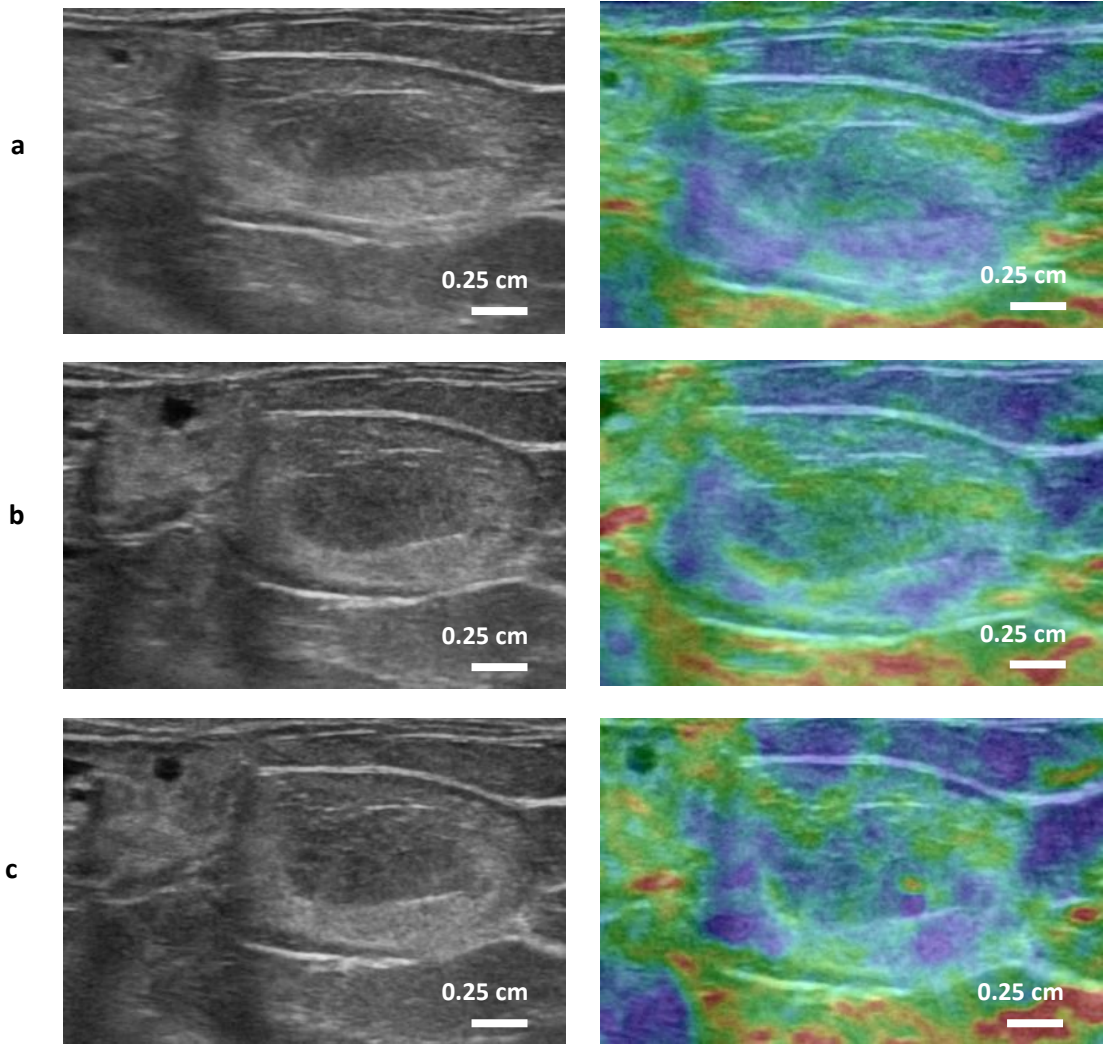

**Supplementary Fig. 1 Multiple B-BUS and SE-BUS images for a patient whose true class is benign.** The proposed ensemble model classifies the images as **(a)** benign, **(b)** benign, and **(c)** malignant.

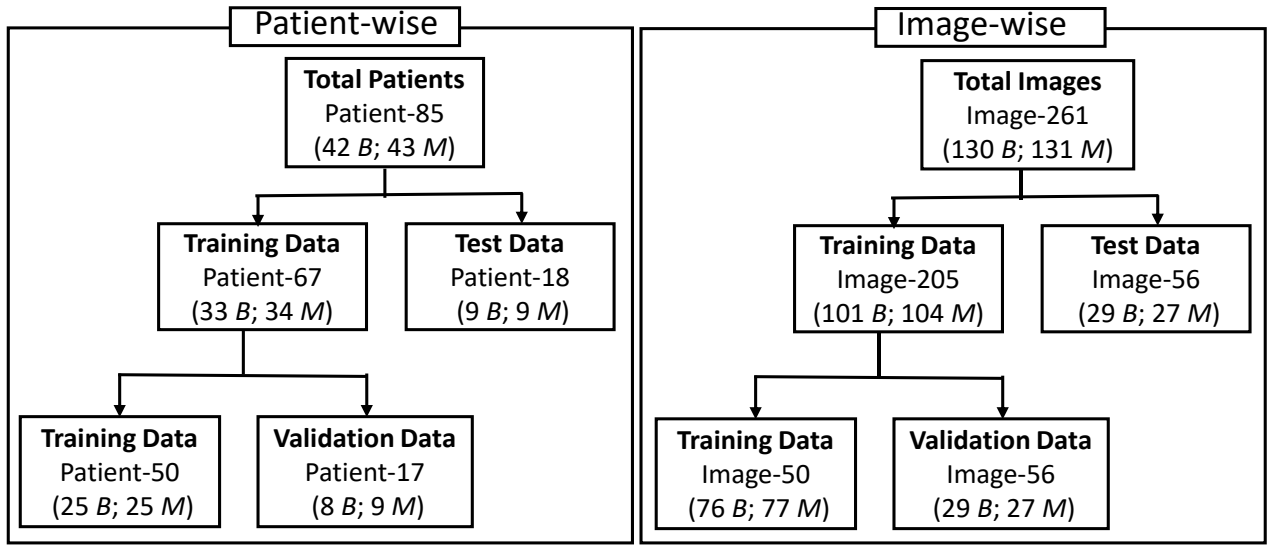

**Supplementary Fig. 2 Patient-wise and image-wise class distribution (train, validation, and test sets).** All images acquired from a specific patient exclusively either belong to test or validation not simultaneously allocating the images to test and validation. *B*, Benign and *M*, Malignant.

**Supplementary Table 1. Details of Benign (*B*) Patients**

| PT # | Type | Strain Ratio | PT # | Type | Strain Ratio |
|---|---|---|---|---|---|
| 224 | benign | 0.96 | 285 | fibroadenoma | 2.42 |
| 290 | benign lymph node | 3.32 | 286 | fibroadenoma | 10.22 |
| 256 | comp cyst | 2.54 | 287 | fibroadenoma | 2.09 |
| 246 | fat necrosis | 5.84 | 291 | fibroadenoma | 2.49 |
| 263 | fat necrosis | 0.20 | 276 | fibrocystic | 2.63 |
| 267 | fat necrosis | 9.66 | 283 | fibrocystic | 1.23 |
| 230 | fibro change | 3.67 | 284 | fibrocystic | 3.33 |
| 231 | fibro change | 1.38 | 249 | fibrocystic change | 5.19 |
| 247 | fibroaden hyperplaia | 0.87 | 252 | fibrocystic change | 2.70 |
| 203 | Fibroadenoma | 1.75 | 277 | fibrosis | 4.57 |
| 211 | fibroadenoma | 6.17 | 279 | fibrosis | 0.75 |
| 228 | fibroadenoma | 3.57 | 273 | hematoma | 0.38 |
| 233 | fibroadenoma | 1.29 | 269 | hyperplasia | 1.37 |
| 234 | fibroadenoma | 1.70 | 207 | intraductal papilloma | 3.81 |
| 243 | fibroadenoma | 1.63 | 257 | intraductal papilloma | 0.72 |
| 244 | fibroadenoma | 1.52 | 208 | Lipoma | 0.81 |
| 251 | fibroadenoma | 4.40 | 232 | phyolles | 0.92 |
| 264 | fibroadenoma | 4.28 | 260 | scar | 4.81 |
| 265 | Fibroadenoma | 2.64 | 237 | stromal hyperplasia | 1.97 |
| 272 | Fibroadenoma | 2.06 | 241 | Unknown | 1.40 |
| 280 | fibroadenoma | 3.80 | 242 | unknown | 1.68 |
| 281 | fibroadenoma | 2.48 | | | |

**Supplementary Table 2. Details of Malignant (*M*) Patients**

| PT # | Type | Strain Ratio | PT # | Type | Strain Ratio |
|---|---|---|---|---|---|
| 202 | Invasive papillary Ca | 1.36 | 274 | IDC | 13.98 |
| 204 | IDC | 2.15 | 275 | IDC | 4.42 |
| 205 | IDC | 3.04 | 245 | IDC | 16.11 |
| 206 | IDC | 7.29 | 258 | IDC | 10.96 |
| 209 | IDC | 14.03 | 268 | IDC | 12.61 |
| 210 | IDC | 6.07 | 271 | IDC | 1.79 |
| 212 | IDC | 5.95 | 219 | IDC and DCIS | 5.87 |
| 213 | IDC | 7.77 | 220 | IDC and DCIS | 4.98 |
| 216 | IDC | 6.35 | 226 | IDC and DCIS | 10.26 |
| 217 | IDC | 3.52 | 239 | IDC and DCIS | 4.44 |
| 218 | IDC | 18.61 | 254 | IDC c DCIS | 7.57 |
| 221 | IDC | 15.78 | 288 | IDC c DCIS | 2.09 |
| 223 | IDC | 7.50 | 289 | IDC micropapillary | 6.82 |
| 225 | IDC | 4.11 | 248 | IDC mucinous | 4.20 |
| 227 | IDC | 3.45 | 259 | IDC -triple - | 2.96 |
| 229 | IDC | 3.79 | 255 | ILC | 13.99 |
| 235 | IDC | 7.62 | 282 | papilloma with minimal IDC | 6.71 |
| 236 | IDC | 4.79 | 261 | signet cell ca | 6.32 |
| 238 | IDC | 12.34 | 262 | signet cell ca | 4.05 |
| 250 | IDC | 3.22 | 214 | tubular Ca | 2.65 |
| 253 | IDC | 6.97 | 240 | Unknown | 2.87 |
| 270 | IDC | 2.59 | | | |